# Q-Sat AI: Machine Learning-Based Decision Support for Data Saturation in Qualitative Studies

Hasan Tutar[1,2], Caner Erden[3,*], Ümit Şentürk[4]

**Abstract**

The determination of sample size in qualitative research has traditionally relied on the subjective and often ambiguous principle of data saturation, which can lead to inconsistencies and threaten methodological rigor. This study introduces a new, systematic model based on machine learning (ML) to make this process more objective. Utilizing a dataset derived from five fundamental qualitative research approaches—namely, Case Study, Grounded Theory, Phenomenology, Narrative Research, and Ethnographic Research—we developed an ensemble learning model. Ten critical parameters, including research scope, information power, and researcher competence, were evaluated using an ordinal scale and used as input features. After thorough preprocessing and outlier removal, multiple ML algorithms were trained and compared. The K-Nearest neighbors (KNN), Gradient Boosting (GB), random forest (RF), XGBoost, and decision tree (DT) algorithms showed the highest explanatory power (Test $R^2 \approx 0.85$), effectively modeling the complex, non-linear relationships involved in qualitative sampling decisions. Feature importance analysis confirmed the vital roles of research design type and information power, providing quantitative validation of key theoretical assumptions in qualitative methodology. The study concludes by proposing a conceptual framework for a web-based computational application designed to serve as a decision support system for qualitative researchers, journal reviewers, and thesis advisors. This model represents a significant step toward standardizing sample size justification, enhancing transparency, and strengthening the epistemological foundation of qualitative inquiry through evidence-based, systematic decision-making.

[1] Bolu Abant Izzet Baysal University, Faculty of Communication, 14030 – Merkez, Bolu, Turkey. e-mail: hasantutar@ibu.edu.tr ORCID: 0000-0001-8383-1464

[2] Affiliation University: Research Methods Application Center, Azerbaijan State University of Economics (UNEC). E-mail: hasantutar@unec.edu.az

[3] Department of Computer Engineering, Faculty of Technology, Sakarya University of Applied Science, Sakarya, Turkey, e-mail: cerden@subu.edu.tr, ORCID: 0000-0002-7311-862X

[*] Corresponding Author

[4] Department of Computer Engineering, Faculty of Engineering, Bolu Abant Izzet Baysal University, 14280, Bolu, Turkey, e-mail: umit.senturk@ibu.edu.tr, ORCID: 0000-0001-9610-9550

## 1. Introduction

Qualitative research methods represent a crucial methodology for gaining an in-depth understanding of the social sciences, human behavior, experiences, and contexts (Bernard, 2013; Dawson, 2015; Patton, 2011). The central premise of this research is to provide a nuanced understanding and interpretation of the complexity inherent in human experiences (Patton, 2015). Qualitative studies contribute to the development of insights into human experiences by providing a deep understanding of participants' contexts, meanings, and perspectives (Bishop, 2006; Dawson, 2015). Researchers further enrich this understanding by examining the social, cultural, and historical factors influencing individuals and groups (Geertz, 1973; Hammersley & Atkinson, 2019). The flexible structure of qualitative research and its adaptability to evolving research questions facilitate contextual understanding of phenomena (Hammersley & Atkinson, 2019; Sandelowski, 1995). These methods are particularly instrumental in comprehending issues such as identity, social interactions, power dynamics, and cultural influences within their respective contexts (Creswell, 1998; Mayring, 2000). Whereas quantitative methodologies primarily address "what" and "how much," qualitative research secures the depth of subjective meaning-making, playing a critical role in elucidating the mechanism and context of phenomena (Geertz, 1973; Moustakas, 1994). This methodological depth often facilitates theory construction by generating new concepts and frameworks (Creswell, 1998). Furthermore, when evaluating the efficacy of policies and programs, qualitative research provides insights beyond mere quantitative outcome measures by exploring the experiences and perspectives of affected individuals (Patton, 2011).

The determination of an appropriate sample size in qualitative research represents a quintessential systemic decision problem characterized by uncertainty, nonlinear interactions among multi-level variables (e.g., research design, data quality, participant heterogeneity), and context-sensitive feedback loops. Traditional reliance on the subjective concept of 'data saturation' fails to adequately model this complexity, leading to arbitrary decisions that compromise scientific consistency. Distinct qualitative research designs capture this depth through varying methodological rationales: Phenomenology aims to investigate the lived experiences and meanings individuals attribute to a specific experience (Moustakas, 1994), providing in-depth descriptions that reveal its essence (Smith et al., 2022). This design yields high information-density texts per observation, generating a strong signal with minimal observations (Hennink & Kaiser, 2022). The case study methodology is valuable for its intensive investigation of a bounded phenomenon, utilizing multiple data sources and its inherent context sensitivity (Creswell & Creswell, 2023; Yin, 2018). Narrative Research examines how individuals construct meaning from their lives and experiences through stories (Clandinin & Connelly, 2000). This design builds experience across the temporal, social, and spatial planes (Clandinin, 2022). Ethnographic Research, conversely, involves the systematic study of a culture or social group through protracted fieldwork in its natural setting (Geertz, 1973; Hammersley & Atkinson, 2019; Laouris & Metcalf, 2025). Each design possesses unique rationales—such as the complexity of the "case," "theoretical saturation," "meaning intensity," and the "breadth of the context"—that directly bear upon sample size determination.

Despite the undeniable value of qualitative research to the body of scientific knowledge, the lack of a definitive methodological standard for determining sample size, often leaving the decision to the researcher's subjective initiative, is widely considered a significant problem (Onwuegbuzie & Leech, 2007). This situation compromises the credibility, consistency, and scientific rigor of the research (Bernard, 2013). Qualitative researchers typically justify their sample size based on the concept of "data saturation," which remains ambiguous and inherently relies on subjective judgment (Charmaz, 2006; Morse, 2015). Although data saturation is often defined as the point at which additional data yields no new information (Francis et al., 2010; O'Reilly & Parker, 2013), this frequently suggests that claims of saturation are efforts at post-hoc rationalization rather than genuine methodological adherence (Onwuegbuzie & Leech, 2007). Even leading experts employ vague and unbounded phrases when discussing sample volume, such as "a well-chosen group," "what will be useful," or "excess" (Lincoln & Guba, 1985; Merriam, 2013; Patton, 2011). The inconsistencies in sample volume recommendations across different qualitative designs further exacerbate this pervasive uncertainty. For example, Creswell (1998) may suggest 5 to 25 interviews for a phenomenological study, whereas Morse (1994) indicates that 15 to 20 participants may be sufficient. Consequently, a demonstrable need exists for a reliable, valid, and highly consistent model for determining sample volume in qualitative research.

To address these significant methodological challenges, this research critically examines the issue of sample volume determination in qualitative studies. It proposes an innovative standard model based on machine learning (ensemble learning), a methodology that has yet to be extensively explored in this context. The core justification for this study lies in addressing methodological uncertainties—such as intuitive or copy-based sample size decisions and the weak saturation resulting from insufficient sampling—by providing a quantitatively evidence-based and objective guidance system that respects the subjective nature of qualitative research (Laouris & Metcalf, 2025; Onwuegbuzie & Leech, 2007). The approach aims to develop a more acceptable and reliable criterion that does not contradict the interpretive nature of qualitative research by utilizing the machine learning (ensemble learning) technique. The model is designed to mitigate the ambiguity of data saturation by incorporating ten critical parameters as input, including research scope, researcher competency, information power, homogeneity versus heterogeneity, data diversity, and interview duration/number. These parameters were established through expert consultation and deliberation. For this purpose, a balanced dataset of examples derived from five core qualitative research designs (Case Study, Grounded Theory, Phenomenology, Narrative Research, Ethnographic Research) was constructed. Preliminary model comparisons established that the Decision Tree Algorithm exhibited the highest explanatory power. The overarching objective is to develop a computational web application that aids qualitative researchers in determining the optimal sample volume. This endeavor aims to establish a methodology that supports advancement in the social and human sciences. This study explicitly contributes to systems-oriented decision architectures by integrating uncertainty, feedback, and design-sensitivity in sample size planning, aligning with the journal's transdisciplinary systems scope.

## 2. Theoretical Framework and Literature Review

The literature on qualitative sample size revolves around three central, interconnected themes: (1) the evolution and refinement of saturation as a guiding principle, (2) the concept of information power as a framework for balancing quality and quantity, and (3) the imperative for design sensitivity in sampling decisions. This review synthesizes these themes to build a theoretical foundation for the systemic, model-based approach proposed in this study. In qualitative research, sample size is a systemic decision problem due to the interaction of multilevel variables and context-sensitive feedback. A systems approach proposes modeling uncertainty, nonlinearity, and design sensitivity simultaneously. This approach requires weighing the concepts of saturation and information power according to the design and context, rather than relying on single-number prescriptions. Recent findings suggest that saturation can be achieved with fewer interviews than previously thought in most contexts. However, this varies depending on the depth of meaning and content, sample heterogeneity, and question coverage (Hennink & Kaiser, 2022). Furthermore, empirical guidelines for minimum sample sizes for different types of qualitative analysis have been developed, providing method-specific decision support for researchers (Wutich et al., 2024). The reflexive thematic analysis literature discusses saturation not as a mechanical threshold but as a decision embedded in the analytical purpose, emphasizing method-design fit (Braun & Clarke, 2021). On the other hand, concept analysis recommends distinguishing between types of saturation, such as code, theme, data, and theoretical saturation, and explicit reporting in sampling decisions (Rahimi, 2024). Recent studies in educational sciences also highlight the misuse of saturation, necessitating contextual justification (Daher, 2023; Schwaninger & Ott, 2025). This framework aligns well with the systemic methodological sensitivity of SRBS and is helpful in theoretical decision-making processes.

In recent years, the distinction between code saturation and semantic/theme saturation has become more pronounced, and this distinction directly impacts sample size decisions in qualitative designs. The saturation threshold varies sensitively to contextual dimensions such as question coverage, data richness, and sample heterogeneity. Therefore, single-number guidelines do not provide a generalizable standard; decisions should be made through design-sensitive assessment. It is reiterated in the literature that sample size can deviate from what is expected depending on the context (Hennink & Kaiser, 2022). Reflexive thematic analysis literature treats saturation not as a mechanical threshold but as a judgment embedded in the analytical purpose (Braun & Clarke, 2021). Recent reviews in educational sciences have highlighted the misuse of saturation and emphasized the need for clear justification (Daher, 2023). Comparative roadmaps across methods provide empirical guidelines on sample size for different types of qualitative analysis, offering researchers context-sensitive options (Wutich, Beresford, & Bernard, 2024). Concept analyses, on the other hand, suggest distinguishing between code, theme, data, and theoretical saturation types, increasing reporting transparency (Rahimi, 2024). Taken together, this evidence suggests that sampling decisions in qualitative studies should be modeled systemically; Current discussions in SRBS and related journals are also moving in this direction. This approach closely supports design-sensitive information power accounts.

The information power approach suggests that sufficient explanation can be produced with smaller samples, given the depth of participant knowledge, sample heterogeneity, and focused research questions. This principle methodologically grounds the quality-quantity premise in qualitative research and makes decision-making processes transparent. Recent syntheses demonstrate that the saturation threshold is context-sensitive and that sample requirements decrease under conditions that increase information power (Hennink & Kaiser, 2022). Recent integrative reviews provide guidelines for five types of saturation and sample size in standard qualitative analysis methods, positioning information power as a weightable input (Wutich, Beresford, & Bernard, 2024). The reflexive thematic analysis literature prioritizes purpose and design congruence over mechanical thresholds, discussing the relationship between information power and interpretative depth (Braun & Clarke, 2021). Concept analysis enhances reporting transparency by distinguishing between code, theme, data, and theoretical forms of saturation (Rahimi, 2024). Studies in educational sciences recommend explicit justification of information power and contextual reporting of sampling decisions (Daher, 2023; Panda, 2024). Thus, the information power approach offers a weightable, transparent, and reproducible framework for system-sensitive decision-support models.

Semantic density in phenomenology, contextual breadth in ethnography, and theoretical saturation in grounded theory studies determine sampling decisions along different trajectories. The saturation threshold varies depending on the scope of the question and the richness of the data; single-number prescriptions lack generalizability (Hennink & Kaiser, 2022). Reflexive thematic analysis literature treats saturation not as a mechanical threshold, but as a judgment embedded in the analytical purpose, and prioritizes pattern fit (Braun & Clarke, 2021). Integrative reviews offer design-specific sampling guidelines for various types of qualitative analysis, providing researchers with context-sensitive decision support (Wutich, Beresford, & Bernard, 2024). Conceptual analyses strengthen reporting transparency by clarifying the distinctions between code, theme, data, and theoretical saturation (Rahimi, 2024). A systems thinking perspective proposes managing complexity through diversity engineering, suggesting a systemic construction of design-sensitive, multivariate decisions (Schwaninger & Ott, 2025). Thus, within the phenomenology-ethnography-theorizing triad, sampling decisions can be optimized within a systemic framework, encompassing variables such as context, purpose, and knowledge strength.

Machine learning and ensemble approaches are superior at capturing nonlinear interactions and threshold effects; therefore, the predictive power that feeds sampling decisions increases (Ekici, Önsel Ekici, Yumurtacı Hüseyinoğlu, & Watson, 2024). In systems-oriented research, transparent reporting of method selection and error metrics together enhances reproducibility; in particular, presenting MAE and RMSE together increases comparability (Collins et al., 2024). For reliable estimation, the conformal prediction family offers uncertainty bands with model- and distribution-independent coverage guarantees, thereby enhancing transparency in decision-making processes (Angelopoulos & Bates, 2023). Comprehensive reviews detail application designs, calibration logic, and performance metrics in classification and regression contexts (Zhou et al., 2026). Conformal approaches, integrated with expert-human judgment, promote sensitive analysis designs that support sampling decisions by increasing perceived accuracy and explainability (Straitouri et al., 2023). This trend enables methodological standardization consistent with systems thinking within the Systems Research and Behavioral Science ecosystem.

Saturation and design sensitivity are discussed extensively in the literature; however, a systemic decision-support model that incorporates multivariate and data/artifact quality appears to be limited. The Systems Research and Behavioral Science strand promotes decision architectures that account for uncertainty and feedback (Schwaninger & Ott, 2025; Zarghami, 2024). This study integrates inputs such as information power, sample heterogeneity, researcher competence, and data diversity within a machine learning-based framework (Panda, 2024). The approach enhances reproducibility by jointly reporting error metrics such as MAE/RMSE. Furthermore, conformal prediction and similar robust estimation techniques increase decision transparency by clearly presenting uncertainty bands. This direction aligns with the SRBS discourse on system dynamics and design-oriented systems applications (Lane, 2024; Sweeting & Sutherland, 2023). It is reiterated in the qualitative methods literature that the saturation threshold is context-sensitive and that "single-number" prescriptions are limited in their applicability. The proposed model quantifies this evidence through design-sensitive weightings, optimizing sampling decisions in conjunction with data quality, question coverage, and analytical objectives. This ensures methodological standardization consistent with systems thinking and makes a concrete contribution to the SRBS methodological innovation agenda with a decision-support focus.

## 2.1. Sampling and Saturation in Qualitative Research

Given that the fundamental philosophy of qualitative research centers on understanding and interpreting the complexity of human experiences, it is evident that sample size cannot be determined with the statistical precision characteristic of quantitative research. Within this context, the principal criterion governing sample volume in qualitative studies has been the concept of theoretical or thematic saturation. The concept of saturation was initially introduced by Glaser and Strauss (1967) as "theoretical saturation" within the Grounded Theory approach. Theoretical saturation denotes the point at which "collecting further data regarding a theoretical construct yields no novel information and ceases to contribute to greater confidence in the grounded theory"; at this juncture, the conceptual categories forming the theory are deemed saturated. The core rationale employed to attain theoretical saturation involves data-driven sampling (theoretical sampling) dedicated to theory construction.

Although the majority of qualitative research does not adhere strictly to the Grounded Theory approach, the concept of saturation has subsequently been adopted across other qualitative methodologies. In a broader application, saturation has shifted its focus from the sufficiency of data for theory development (as defined in theoretical saturation) to the generalized assessment of sample size adequacy. This broader concept is referred to as "data saturation." Data saturation refers to the condition in which no further topics or insights are uncovered during data collection, and the data exhibit redundancy (Francis et al., 2010; Morse, 2015; O'Reilly & Parker, 2013). It is assumed that the collected data sufficiently capture the range, depth, and nuances of the phenomena under study. Later researchers, addressing the intrinsic ambiguity of data saturation, have refined the concept into more specific typologies: Code Saturation (the cessation of novel codes or thematic fragments emerging) and Meaning/Thematic Saturation (defined as the point where the accumulated data adds no substantive new information to the research). For example, Hennink et al. (2022) suggest that meaning saturation can typically be achieved with approximately 24 interviews.

The primary distinction between these saturation approaches lies in their objective: Theoretical saturation is intrinsically linked to a theory-building process, while data/thematic saturation is employed to justify the sufficiency of the sample size. Nevertheless, the reliance of data saturation on entirely subjective judgment (Bernard, 2000; Charmaz, 2006; Sweeting & Sutherland, 2023) diminishes the credibility and consistency of qualitative research. Unsupported assertions of reaching saturation weaken the concept's methodological value. This ambiguity has led to a broad and inconsistent spectrum of documented sample sizes in the literature: Bernard (2000) recommends 30 to 50 participants, while Charmaz (2003) suggests 25-30. Guest et al. (2006) propose 30–60 interviews for ethnographic studies, yet indicate that 6–12 interviews might suffice for a general qualitative project. Mason (2010) documented a range of 1 to 95 cases in case studies and 1 to 62 (life stories) in narrative research. The strategy or design of qualitative research (such as case study, phenomenological research, ethnographic research, grounded theory, and narrative research) is a direct determinant of sample size. Each research design possesses a unique objective for saturation (Table 1).

Table 1. Sample Size Ranges and Saturation Approaches According to Qualitative Research Designs

| Research Design | Core Requirement and Saturation Goal | Sample Size References |
|---|---|---|
| Phenomenology | Capturing the Essence of Lived Experiences: Aims to investigate the lived experiences and meanings individuals attribute to a specific experience. The goal is to provide in-depth descriptions that reveal the essence of the experience. This design conceptualizes saturation through depth of meaning and operates with small, typically homogeneous samples. | Creswell (1998) 5–25 interviews; Morse (2020) 15–20 participants. A wide range of 3 to 372 exists in the literature. |
| Narrative Research | In-Depth Exploration of a Single Life Story: Examines how individuals construct meaning from their lives and experiences through narratives. Sample selection requires strategies that preserve the integrity and depth of the narrative. | Generally focuses on 2 or 3 cases, as suggested by Creswell (2023). Studies in the literature have included a minimum of 1 and a maximum of 143 participants. |
| Ethnographic Research | Comprehensive Account of Cultural Context: Involves the systematic study of a culture or social group through extended fieldwork in its natural setting. The objective is to capture cultural patterns, interactions, and practical knowledge; the unit of analysis is frequently events and networks of relationships. Based on contextual understanding, sampling predominantly utilizes "purposive" and "snowball" techniques. | Morse (1994) suggested 30–50 interviews, while a wide range of 1 to 160 is found in the literature. |
| Case Study | Detailed Analysis of a Bounded Phenomenon: It is crucial for the intensive investigation of a bounded phenomenon using multiple data sources, thereby transferring context sensitivity to the data universe. The objective is to gain a deep understanding of the complexities, dynamics, and unique characteristics of the selected case. | Creswell (2023) suggests 4 to 5 cases. Yıldırım and Şimşek (2016) state that the number of individuals included in the sample should generally not exceed 10. However, this design is observed to exhibit the most pronounced positive skewness in practice, with extreme outliers extending to over 300 units. |

Each of these designs possesses distinct rationales that directly influence sample size decisions: "meaning intensity" in Phenomenology, the complexity of the "case" in Case Study, "theoretical saturation" in Grounded Theory, and "breadth of the context" in Ethnography. In qualitative research, the intrinsic quality of the data obtained per sampling unit holds greater significance than numerical quantity. Nonetheless, this emphasis on quality often leads researchers toward highly subjective judgments, perpetuating the fundamental problem of saturation, as it lacks standardized criteria.

## 2.2. Factors Influencing Sample Size

Sample size determination in qualitative research has traditionally been governed by the subjective concept of "data saturation." To address this fundamental problem and introduce a quantifiable standard for sample volume, the machine learning-based model developed utilizes ten critical input parameters (features) designed to balance the qualitative depth and breadth dilemma. These parameters facilitate the estimation of the optimum sample volume by reflecting the research's data quality, richness, and context-specific requirements. The ten core features utilized in the model are encoded on a three-level ordinal scale (with scores of 15, 20, or 25, or 10, 15, 20 points); as shown in Table 2, each exerts a direct theoretical influence on the sample size decision.

Table 2. Fundamental Factors and Their Theoretical Effects

| Factor | Theoretical Impact and Literature Relationship | Relationship to Sample Size |
|---|---|---|
| Research Scope | Depth vs. Breadth: A broader research objective necessitates a larger sample size than a narrow one to ensure adequate information power. Greater scope implies the need for a larger and more diverse sample volume to capture the phenomenon from multiple perspectives. Morse (2000) and Patton (2015) emphasize that the clarity of the objective is a key determinant of sample size. | Broad-scope studies require more participants (25), whereas narrow-scope research requires fewer participants (15). |
| Information Power | Quality vs. Quantity: This concept, proposed by Malterud, Siersma, and Guassora (2016), suggests that increased participant knowledge about the research topic enables a smaller sample volume. High information power implies that theoretical saturation is attainable even with a minimal number of participants. Information power correlates with the study's capacity to elucidate its objectives (Kuzel, 1999; Morse, 1994; Sandelowski, 1995). | Expert participants (high information power) require fewer participants (15); conversely, less informed participants (low information power) necessitate a larger group (25) to compensate for the information deficit. |
| Homogeneity/Heterogeneity | Balance of Diversity and Depth: A greater variation in participant characteristics or experiences (heterogeneity) necessitates a larger sample size to ensure comprehensive capture of diverse perspectives. Conversely, saturation is theorized to be achieved earlier in homogeneous samples (Glaser & Strauss, 1967). The imperative for heterogeneity underscores the importance of originality and novel information over assumed uniformity. | A heterogeneous participant group mandates more participants (25), while a homogeneous participant group requires fewer participants (15). |
| Researcher Competence | Quality of Interpretation: A researcher's competence in interpretation reflects their capacity to comprehend collected data, identify underlying patterns, and derive conclusions efficiently. High competency offers the potential to extract high-quality information from fewer participants, enabling researchers to effectively interpret nuances and more profound meanings. | A highly competent/expert researcher can operate with a relatively smaller sample (15); a less competent researcher may be constrained to utilize a larger sample (25) (a broader data pool). |
| Number and Duration of Interviews | Intersection of Depth and Breadth: A higher number of interviews conducted per participant generally permits a smaller sample volume, as the total volume of collected data increases. Similarly, a longer allocated interview duration yields a greater data volume, potentially reducing the need for more participants (Marshall et al., 2013; Morse, 2000). | If the number of interviews exceeds five, 15 participants may suffice; if the number is less than five, a larger sample (25) is suggested to mitigate the information deficit. If the interview duration exceeds 2 hours, 15 participants are sufficient; for durations between 1 and 2 hours, 25 participants are required. |
| Participant Originality | Unique Information Production: Selecting highly original and knowledgeable participants in purposive sampling ensures a high density of data collected from a limited number of individuals. Originality can either delay or potentially obviate the need for data repetition ("data saturation"), adhering to the principle that novel information is always present within a heterogeneous group. | Work is feasible with participants possessing a high capacity for generating new information (15), whereas participants with low capacity necessitate a larger sample (25). |
| Data Diversity (Triangulation) | Reliability and Scope: Employing multiple data sources within the same research (e.g., observation, interviews, documentary sources) enhances the methodological foundation and mitigates the risk of "systematic error". Utilizing diverse data sources increases the depth and scope of information, thereby lessening the necessity to compensate for an information gap. | If data diversification is high (resulting in high reliability), a smaller sample size (15 participants) may be sufficient; if diversification is low, a larger sample size (25 participants) is required. |
| Data Quality | Sufficiency: This metric is determined by the competence of the participants providing data and the data's potential to generate codes, categories, and themes (Kvale, 1996; Sandelowski, 1995). A sufficient data volume can be reliably secured from a relatively small number of high-quality participants (Morse, 2000; Patton, 2015). | A smaller sample (15) suffices for high-quality data, while a larger sample (25) is recommended to compensate for deficiencies in low-quality data. |

**Emphasis on Depth vs. Breadth Decision:** The core mandate of qualitative research emphasizes quality over quantity. In this vein, Patton (2015) posits that even a single participant can be adequate provided they satisfy the requirements of purposive sampling. Sandelowski (1995) argues that huge sample sizes preclude the possibility of truly in-depth interviews. Metrics such as Information Power, Participant Originality, Data Quality, and Interview Duration/Number quantify this emphasis on depth (quality). High scores on these metrics (e.g., Information Power score of 15) consequently tend to reduce the estimated sample size, reflecting the assumption that sufficient quality negates the need for excessive quantity.

Conversely, metrics including Research Scope, Homogeneity/Heterogeneity, and Research Strategy/Design (e.g., Ethnographic and Grounded Theory approaches) quantify the emphasis on breadth (scope). High scores on these metrics (e.g., Heterogeneity score of 25) tend to increase the estimated sample size. The model is meticulously designed to identify the optimal point between these opposing assumptions: determining a sample size that is both adequate to avoid an information deficit and restrained enough to permit rigorous, in-depth analysis. Each of the ten features incorporated into the model is structured within the context of this depth-breadth dichotomy, facilitating a standardized and objective decision-making process free from undue subjective influence.

## 3. Methodology

### 3.1. Data Set and Data Preprocessing

The primary objective of this research was to establish a reliable, balanced, and comprehensive dataset suitable for analyzing sample size determination processes in qualitative research. The dataset creation phase, which encompassed the systematic collection, preparation, and structuring of data for the training, validation, and testing of the machine learning model, constitutes the foundational element of this study. A systematic literature search yielded the screening of 730 qualitative research articles across various databases, including TR-Dizin, Scopus, SSCI, SSCI-E, and ESCI. The study systematically analyzed five widely adopted qualitative research designs representing different epistemological orientations. The initial raw dataset comprised studies in the following categories: Case Study (219), Grounded Theory (151), Phenomenology (151), Narrative Research (120), and Ethnographic Research (80).

The selected academic works (articles and theses) underwent meticulous analysis. This rigorous process involved scoring each qualitative research article based on ten predetermined metrics. This multi-stage analysis utilized the independent assessments and scoring of three distinct experts to ensure methodological objectivity and reliability. Following data cleaning, validation, and equalization procedures, data imbalance was addressed by ensuring an equal distribution across each research design for model training. The dependent variable (Sample Size) represents the actual sample volume (number of participants/cases) reported in each qualitative research article included in the analysis. Ten distinct metrics reflecting the quality and data richness of qualitative research were utilized as input features for the machine learning model. These features are intended to establish a standard criterion free from excessive subjective judgment. These metrics were encoded on a three-level ordinal scale determined through expert consultation. As shown in Table 3, the scale range is 10–20 points, with values representing scores of 10 (Low), 15 (Medium), and 20 (High).

This scoring system was developed based on the philosophical underpinnings of qualitative research, aiming to provide a standard criterion free from excessive subjective judgments. In qualitative research, the intrinsic quality of the data obtained per sampling unit is paramount. Thus, the scoring methodology reflects:

**Emphasis on Quality:** Metrics such as Information Power, Participant Originality, and Data Quality tend to reduce the sample size when assigned high scores (e.g., 15 points). This confirms the assumption that sufficient quality (in-depth, rich information) obviates the need for excessive quantity, aligning with the qualitative methodological emphasis of Sandelowski (1995) and Patton (2014).

**Scope and Difficulty:** Metrics such as research scope and Homogeneity/Heterogeneity quantify the difficulty and the required breadth of the research. For instance, a heterogeneous group (25 points) necessitates more participants to capture diverse perspectives.

Table 3. Independent variables and scoring examples

| | **Precise Definition and Theoretical Effect** | **Concrete Scoring Example** |
|---|---|---|
| Data Quality | Related to ensuring data integrity (obtained from competent participants) and the data's generative capacity for codes, categories, and themes. Data quality is directly correlated with the quality of both participants and interviewers. A smaller sample suffices for high-quality data, whereas an information deficit must be compensated for when data quality is low. | Low-quality data is assigned 25 points (or 10 points), medium-quality data 20 points, and high-quality data 15 points (or 25 points). A higher quality score resulting in a smaller sample (15 points) reflects the priority given to quality. |
| Information Power | Pertains to the depth of participants' knowledge regarding the research topic. High information power allows theoretical saturation to be achieved even with a relatively small sample (Malterud et al., 2016). | If the participant is an expert in the research topic, 15 points are assigned; if they are knowledgeable to a moderate extent, 20 points are assigned; if they are not sufficiently knowledgeable, 25 points are assigned to the hypothetical participant group. |
| Homogeneity | Expresses the diversity of the sample's characteristics or experiences (heterogeneity). Increased heterogeneity necessitates a larger sample volume (25 points) to capture diverse perspectives. | If operating with a homogeneous participant group, 15 points are assigned; if operating with a heterogeneous participant group, 25 points are assigned for determining the optimal number. |
| Number of Interviews | A higher number of interviews conducted per participant permits a smaller sample volume (15 points), as the data volume collected per person increases. | If the number of interviews exceeds five, 15 points are assigned; if the number of interviews is less than five, 25 points are assigned for the hypothetical participant. |
| Researcher Competence | Quality of Interpretation: Refers to the researcher's ability to interpret collected data, identify patterns, and draw robust conclusions efficiently. High competency (expertise) offers the potential to extract high-quality information from fewer participants. | A highly competent/expert researcher can operate with a relatively smaller sample (15); a less competent researcher may be constrained to work with a larger sample (25) (a broader data pool). |
| Research Scope | Depth vs. Breadth: A broader research objective requires a larger sample size than a narrow objective to ensure adequate information power. A greater scope implies that a larger, more diverse sample volume will be needed to comprehend the phenomenon from multiple perspectives. Morse (2000) and Patton (2015) emphasize that objective clarity is a key determinant of sample size. | Broad-scope studies typically require more participants (25), whereas narrow-scope research generally requires fewer participants (15). |
| Data Diversity (Triangulation) | Reliability and Scope: Utilizing multiple data sources (observation, interview, documentary sources) within the same research enhances the methodological foundation and mitigates the risk of "systematic error" (Maxwell, 1996). Data diversification increases the depth and scope of information obtained, thereby reducing the necessity to compensate for an information deficit. | If data diversification is high (resulting in high reliability), a smaller sample size (15 participants) may be sufficient; if diversification is low, a larger sample size (25 participants) is required. |
| Participant Originality | Unique Information Production: Selecting original and knowledgeable participants in purposive sampling ensures a high density of information collected from a limited number of individuals. Originality can delay or potentially eliminate the point of data repetition ("data saturation"). | Work is feasible with participants capable of providing high amounts of new information (15), whereas participants with low capacity necessitate a larger sample (25). |
| Interview Duration | Intersection of Depth and Breadth: A longer allocated interview duration results in greater data volume, potentially reducing the need for more participants. | If the interview duration exceeds 2 hours, 15 participants are sufficient; for durations between 0 and 1 hour, 25 participants are required. |

These metrics collectively aim to prevent arbitrariness in determining sample size by providing objective input to the machine learning model regarding thematic richness, redundancy, and diversity of perspectives.

The data preprocessing stage was executed after the dataset creation to ensure the reliability and generalizability of the machine learning model. This stage involved critical steps to prepare the collected data for the subsequent modeling phases and to enhance the dataset's capability to generate meaningful information, particularly for ensemble learning algorithms. The distribution of the sample size exhibits a pronounced positive skewness, evidenced by the mean exceeding the median. This distribution indicates that a small number of studies with large values extend the tail, posing a risk of impact/leverage from extreme outliers. The 95th percentile approach was utilized for outlier detection, effectively removing statistically deviant data points. This trimming procedure successfully reduced variance, diminished skewness, and stabilized the central tendency measures. For example, the mean of the Grounded Theory design decreased from 41.0 to 26.7 following outlier trimming, demonstrating successful management of the influence of extreme upper-tail demands. Figure 1 illustrates the distribution of sample sizes within each qualitative research design, including *Grounded Theory, Phenomenology, Narrative Research, Ethnographic Research,* and *Case Study*. Figure 2 represents the final dataset distributions.

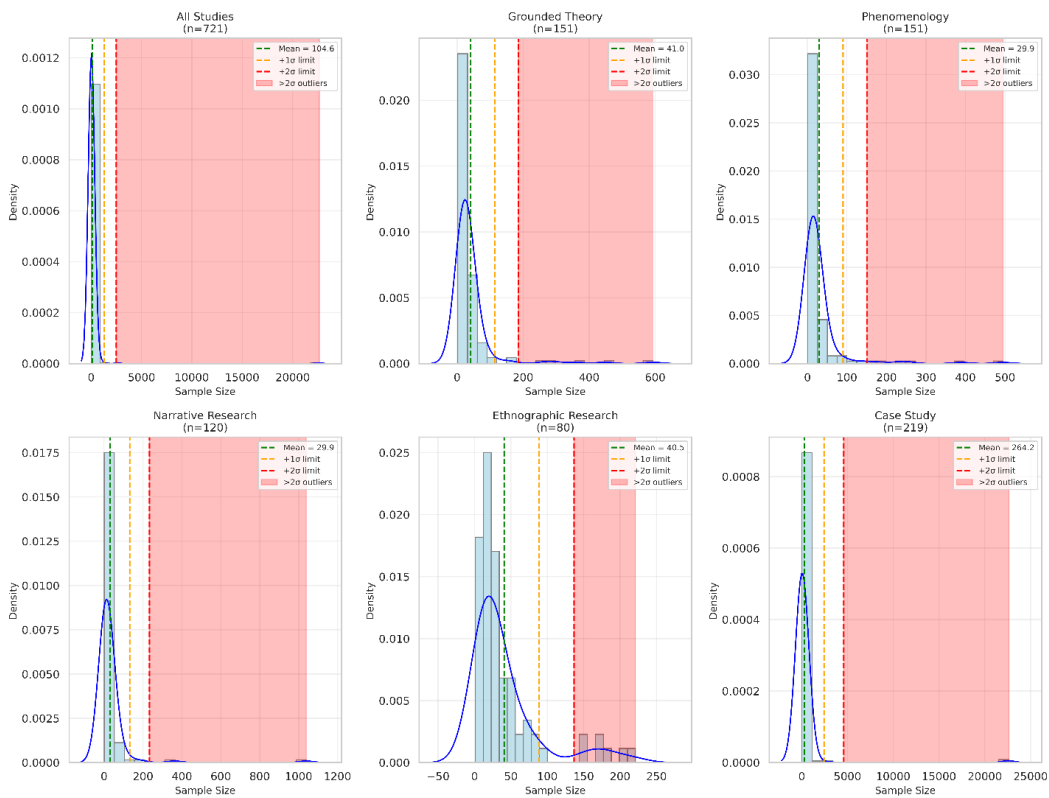

Figure 1. Outlier Analysis of Sample Sizes Across Qualitative Research Designs Using the Standard Deviation Rule

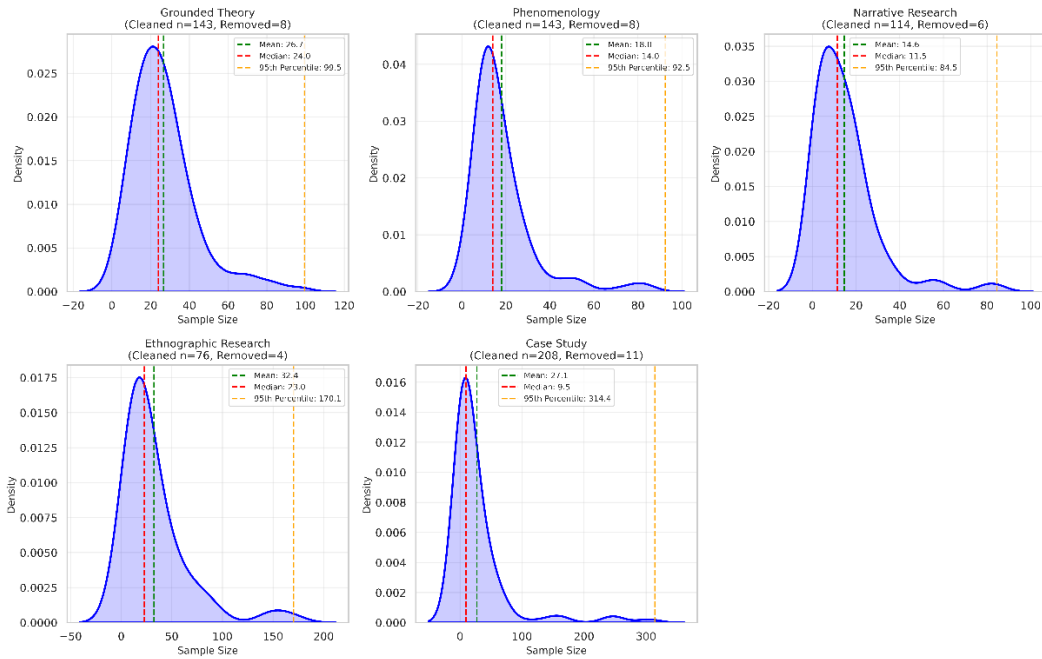

Figure 2. Distributions of the dataset after the data preprocessing steps

The conspicuous positive skewness of the target variable necessitated transformation to mitigate potential violations of normality and homoscedasticity assumptions inherent in linear models. The applied transformation, log-transformation (Log(y)), facilitated enhanced adherence to model assumptions. This technique is crucial as the positive tail and excessive data scatter weaken the Poisson assumption, enabling the model to generate more accurate and generalizable predictions.

The Research Strategy/Design (e.g., Grounded Theory, Phenomenology) serves as a categorical variable input. Categorical variables were converted via numerical encoding. Inconsistencies arising from varying nomenclature were meticulously resolved through code merging and standardization. This ensured the uniform representation of categories, mitigating the risk of erroneous learning during model training.

Ten metrics, each encoded on an ordinal scale (10, 15, 20), were utilized as independent variables. All numerical features underwent normalization using a standardized scaler. Scaling is mandatory for distance-based (e.g., Support Vector Machines (SVM)) and gradient-based algorithms (e.g., Gradient Boosting (GB), as well as Artifivial Neural Networks (ANN). Normalization ensures that features across different scales do not disproportionately influence the model, thereby enhancing the consistency of predictions.

Consequences of Skipping Scaling: Had the scaling step been omitted:

- Algorithms such as ANN would have disproportionately weighted features with inherently larger scales, effectively negating the influence of smaller-scale features.
- Training convergence for algorithms (especially ANN) would have been slower and more unstable, as gradient descent optimization would struggle to reach the optimal solution.
- Error metrics (RMSE, MAE) would have increased significantly, rendering predictions unreliable.

Consequently, the meticulous data preprocessing pipeline was indispensable for ensuring the successful operation of the machine learning model on this dataset, given the complex, non-linear patterns and high inherent variance of qualitative research data.

### 3.2. Model Architecture and Training

We pre-registered hyperparameter grids for tree ensembles and MLP, reported selection criteria (MAE-first, RMSE-second), and retained 5-fold OOF predictions for the meta-learner. The Stacking Regressor was selected as the final ensemble method based on a core principle of ensemble learning: to leverage the diverse strengths and complementary error patterns of multiple base models. While a single model, such as a Decision Tree, might achieve high performance, a well-designed stack can yield a more robust and generalizable predictor by synthesizing these varied perspectives, thus mitigating the risk of over-reliance on any single model's inductive bias. The machine learning model developed for estimating the qualitative sample size is based on the Ensemble Learning method, strategically designed to leverage the complementary strengths of multiple distinct algorithms. This prediction process aims to accurately model the complex, non-linear relationships existing between the independent variables (ten metrics) and the continuous dependent variable (sample size).

Fundamental algorithms were developed for sample size prediction:

- K-Nearest Neighbors (KNN): KNN is a non-parametric, instance-based learning algorithm that classifies data points based on the majority label of their nearest neighbors in the feature space. It is particularly effective in low-dimensional datasets and requires no prior assumptions about data distribution (Halder et al., 2024).
- Gradient Boosting (GB): GB is an ensemble method that builds models sequentially, where each new model attempts to correct the errors of its predecessor. It uses gradient descent optimization to minimize loss and is known for achieving high predictive accuracy, especially in structured data (Bentéjac et al., 2021).
- Random Forest (RF): RF is a bagging-based ensemble technique that constructs multiple decision trees using random subsets of features and data. It reduces overfitting and improves generalization performance, making it robust for both classification and regression tasks (Breiman, 2001).
- eXtreme Gradient Boosting (XGBoost): XGBoost is an optimized implementation of gradient boosting that incorporates regularization, parallel processing, and sparse-aware learning. It is widely used in machine learning competitions due to its scalability and efficiency (Chen & Guestrin, 2016).
- Decision Tree (DT): DTs are rule-based models that recursively split data based on feature values to form a tree structure. They are interpretable and intuitive, often serving as base learners in ensemble methods like RF and GB (Blockeel et al., 2023).
- Support Vector Regression (SVR): SVR extends Support Vector Machines to regression tasks by fitting a function within a margin of tolerance (ε-insensitive loss). It is effective in modeling non-linear relationships in high-dimensional spaces using kernel functions (Awad & Khanna, 2015).
- Multi-Layer Perceptron (MLP): MLP is a feedforward artificial neural network with one or more hidden layers. It uses backpropagation for training and can model complex, non-linear relationships. MLPs are foundational in deep learning architectures (Przybyła-Kasperek & Marfo, 2024).

- Adaptive Boosting (AdaBoost): AdaBoost combines multiple weak learners by iteratively adjusting their weights based on classification errors. It focuses on difficult samples and is known for its low bias and strong generalization in binary classification (Beja-Battais, 2023).
- Ridge Regression: Ridge Regression introduces L2 regularization to linear regression, penalizing large coefficients to reduce variance. It is particularly useful when predictors are highly correlated, improving model stability and interpretability (Shalabh, 2022).

The 5-fold cross-validation (CV) method was implemented throughout the model development and validation processes. This strategy ensures a reasonable trade-off between bias and variance, particularly relevant given the dataset's scale. To prevent overfitting and enhance the model's generalization ability within the Stacking architecture, Out-of-Fold (OOF) predictions are generated. OOF predictions represent the estimates made by each base model on the data partitions it has never seen during its training phase. The OOF predictions from the five base models are concatenated to construct a new input dataset. This resultant dataset constitutes the input layer for the Meta-Model (e.g., Linear Regression or Elastic Net). The Meta-Model processes these base model outputs to deliver the final prediction of the sample size. This architectural design integrates the predictions from highly capable tree-based composites and MLP—both adept at capturing complex, non-linear relationships inherent in qualitative data—thereby maximizing prediction consistency.

## 4. Findings

### 4.1. Descriptive Statistics

The descriptive statistics characterizing the dataset utilized for qualitative sample size estimation are presented in this section. These foundational statistics were crucial for understanding the intrinsic data characteristics and informing the selection of suitable machine learning approaches during the model development process.

Analysis of the target variable's distribution reveals a pronounced positive skewness. Table 4 presents the central tendency and dispersion measures for each independent variable, including the mean, standard deviation, quartiles (Q1–Q3), and distribution shape indicators (skewness and kurtosis). The results demonstrate variability among methodological quality indicators and sample size, reflecting diverse study designs and data collection characteristics.

Table 4. Descriptive statistics of independent variables used in the analysis

|  | Mean | Std. Dev. | Min | Q1 | Median | Q3 | Max | Skewness | Kurtosis |
|---|---|---|---|---|---|---|---|---|---|
| **Qualitative Design Score** | 19.29 | 6.06 | 7.5 | 17.5 | 22.5 | 25 | 25 | -0.93 | -0.35 |
| **Research Scope** | 17.31 | 5.11 | 10 | 15 | 20 | 20 | 115 | 9.39 | 183.81 |
| **Researcher Competence** | 18.86 | 2.29 | 10 | 20 | 20 | 20 | 20 | -1.8 | 2.34 |
| **Knowledge Power** | 13.15 | 3.33 | 10 | 10 | 15 | 15 | 20 | 0.59 | -0.7 |
| **Interview Count** | 11.26 | 2.96 | 10 | 10 | 10 | 10 | 20 | 2.21 | 3.44 |
| **Interview Duration** | 14.53 | 4.54 | 10 | 10 | 15 | 20 | 20 | 0.19 | -1.76 |
| **Observation Duration** | 11.82 | 3.68 | 10 | 10 | 10 | 10 | 20 | 1.64 | 0.85 |
| **Homogeneity** | 15.2 | 4.24 | 10 | 10 | 15 | 20 | 20 | -0.08 | -1.6 |
| **Participant Originality** | 10.85 | 2.14 | 10 | 10 | 10 | 10 | 20 | 2.51 | 5.79 |
| **Data Variety** | 14.74 | 4.16 | 10 | 10 | 15 | 20 | 20 | 0.1 | -1.55 |
| **Data Quality** | 11.89 | 2.77 | 10 | 10 | 10 | 15 | 20 | 1.13 | 0.28 |
| **Sample Size** | 104.55 | 1193.3 | 1 | 8 | 16 | 31 | 22647 | 18.57 | 346.7 |

## 4.2. Sample Size Comparisons by Research Design

Analyses confirmed differences in sample size distributions across the five core qualitative research designs, as summarized in Table 5.

Table 5. Comparison of Sample Sizes Across Research Designs

| Research Design | Mean | Median | Key Trend |
|---|---|---|---|
| **Ethnographic Research** | 32.4 | 20.0 | Exhibits the highest mean, indicative of a long tail resulting from extended fieldwork requirements. |
| **Grounded Theory** | 26.7 | 25.0 | The mean and median demonstrate proximity; the distribution's tail is short, and the data density clusters around the 20–30 range. |
| **Case Study** | 27.1 | 10.0 | Shows the highest skewness. The contrast between a low median (10.0) and a high mean (27.1) suggests most cases employ small samples, though a few large-scale studies significantly extend the tail. |
| **Phenomenology** | 18.0 | 13.5 | Distribution is narrow and positively skewed, peaking in the 10–20 range, reflecting the rationale of in-depth analysis with small, homogeneous samples. |
| **Narrative Research** | 14.6 | 12.0 | The mean and median are low; the data density peaks in the 10–15 range. |

Evidently, Ethnographic Research (32.4) and Grounded Theory (26.7) utilize larger average sample volumes compared to Phenomenology (18.0) and Narrative Research (14.6) designs. The elevated mean in the Case Study, despite its low median, underscores the presence of methodological outliers.

Outlier trimming effectively managed the influence of extreme values. For Grounded Theory, the mean decreased from 41.0 to 26.7; for Phenomenology, from 28.9 to 18.0; and for Narrative Research, from 23.9 to 14.6, demonstrating increased convergence toward the median. The application of Logarithmic Transformation (Log(Sample Size)) proved essential for stabilizing the positive skewness, enhancing the model's adherence to statistical assumptions. These trends validate the qualitative research practice dichotomy: small, homogeneous studies tend to prefer smaller samples. At the same time, approaches focused on contextual and theoretical development (such as ethnography and grounded theory) necessitate larger samples. This confirms that a universally applicable single-sample rule is not optimal, and the prediction model must be sensitive to the chosen research design.

## 4.3. Model Performance Evaluation

The performance results of the five fundamental machine learning algorithms developed for optimum sample size estimation, alongside the Stacking Regressor ensemble model, are presented in Table 6. Model performance was systematically compared using the coefficient $R^2$ RMSE (Root Mean Squared Error), and MAE (Mean Absolute Error) metrics. As demonstrated in Table 6, the KNN, GBRF, XGBoost, and DecisionTree algorithms achieved the highest goodness-of-fit, explaining approximately 85% of the variance in the target variable. The SVR algorithms exhibited highly competitive performance. Notably, the significantly lower $R^2$ values associated with simple linear models (Ridge and Lasso) suggest that the relationships within the dataset are distinctly non-linear, highlighting the importance of complex feature interactions. MAE serves as a crucial metric for evaluating absolute error, where minimal values are optimal. The MAE value for the best-performing DT is 14.74. This metric quantifies the average number of participants/units by which the model's estimate deviates from the actual sample size.

Table 6. Comparative Performance of Machine Learning Models

| Model | Test R² | Train R² | Test MAE | Best Params |
|---|---|---|---|---|
| KNN | 0.852742 | 0.909446 | 0.151285 | {'model__n_neighbors': 15, 'model__p': 1, 'model__weights': 'distance'} |
| GB | 0.852534 | 0.907133 | 0.17568 | {'learning_rate': 0.1, 'loss': 'squared_error', 'max_depth': 7, 'n_estimators': 200, 'subsample': 0.8} |
| RF | 0.852449 | 0.905714 | 0.183468 | {'max_depth': None, 'max_features': 'sqrt', 'min_samples_leaf': 1, 'min_samples_split': 2, 'n_estimators': 200} |
| XGBoost | 0.849898 | 0.904114 | 0.186369 | {'colsample_bytree': 1.0, 'learning_rate': 0.1, 'max_depth': 7, 'n_estimators': 200, 'reg_alpha': 0, 'reg_lambda': 1.5, 'subsample': 0.8} |
| DT | 0.845724 | 0.91225 | 0.147432 | {'criterion': 'squared_error', 'max_depth': None, 'max_features': 'sqrt', 'min_samples_leaf': 1, 'min_samples_split': 2} |
| SVR | 0.763296 | 0.849767 | 0.263101 | {'model__C': 10.0, 'model__degree': 2, 'model__gamma': 'scale', 'model__kernel': 'rbf'} |
| MLP | 0.685608 | 0.779126 | 0.370804 | {'activation': 'logistic', 'alpha': 0.01, 'early_stopping': True, 'hidden_layer_sizes': (30,), 'learning_rate': 'constant', 'solver': 'lbfgs'} |
| AdaBoost | 0.423281 | 0.438722 | 0.586717 | {'learning_rate': 0.05, 'loss': 'square', 'n_estimators': 100} |
| Ridge | 0.391545 | 0.400687 | 0.57525 | {'model__alpha': 50.0} |

The determination of features that contribute most significantly to the model's predictions provides an empirical foundation for qualitative methodology discussions by objectively quantifying the influence of different qualitative quality metrics on the developed machine learning model. Analysis of feature importance indicates that the variables contributing most significantly to the model's predictions are: The prominence of Research Design Type among the highest-contributing metrics confirms that the model accurately weights the distinct sampling rationales mandated by different qualitative designs (e.g., "meaning intensity" vs. "breadth of context").

The comprehensive performance analysis of the nine base models is visualized in Figure 3. The evaluation of the test set reveals a cluster of high-performing algorithms. The top-tier models—specifically KNN, RF, XGBoost, DT, and GB—all achieved a robust Coefficient of Determination ($R^2$) of approximately 0.85. This finding is highly significant, indicating that these models account for 85% of the variance in the target variable. This explanatory power is exceptionally high, given the inherent complexity of the domain (Sample Size Standard Deviation: 29.3) and the positive skewness of the distribution.

While the $R^2$ scores identify a group of effective models, the Mean Absolute Error (MAE) metric provides a clearer distinction for optimal performance. The DT algorithm yielded the lowest MAE, indicating the most accurate predictions in terms of absolute deviation and positioning it as the best-performing base model overall. Furthermore, Figure 3 was used to analyze model generalization and the risk of overfitting by comparing Train and Test $R^2$. The minimal gap between training and testing performance for the high-performing models (e.g., KNN, DT) confirms that they generalize well to unseen data and are not merely memorizing the training set. Conversely, the poor performance of the simpler linear models (Ridge, AdaBoost), quantitatively corroborates the theoretical assumption that the relationships within the qualitative dataset are distinctly non-linear and rely on complex feature interactions. The model successfully demonstrates the potential to address this complexity using advanced artificial intelligence techniques.

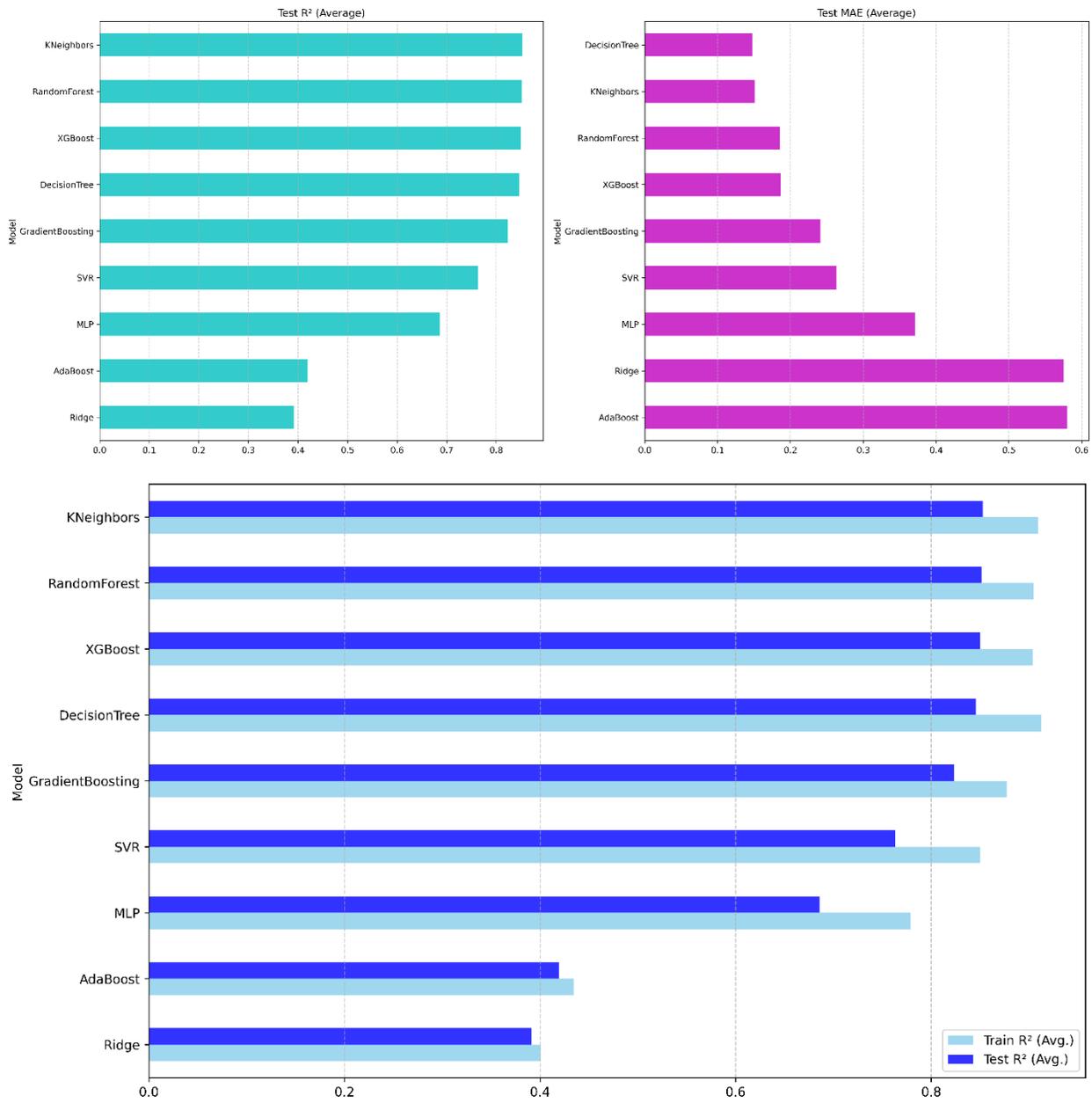

Figure 3. Comparative Model Performance on the test and train set.

The aggregated metrics in Figure 4 reveal a clear distinction in model efficacy. As shown in Figure 4, a cluster of high-performing algorithms—specifically KNN, RF, XGBoost, DT, and GB—all achieved a robust Coefficient of Determination of approximately 0.85, explaining 85% of the variance in the target variable. While the scores are similar, the MAE metric identifies the DT algorithm as the top-performing model with the lowest prediction error.

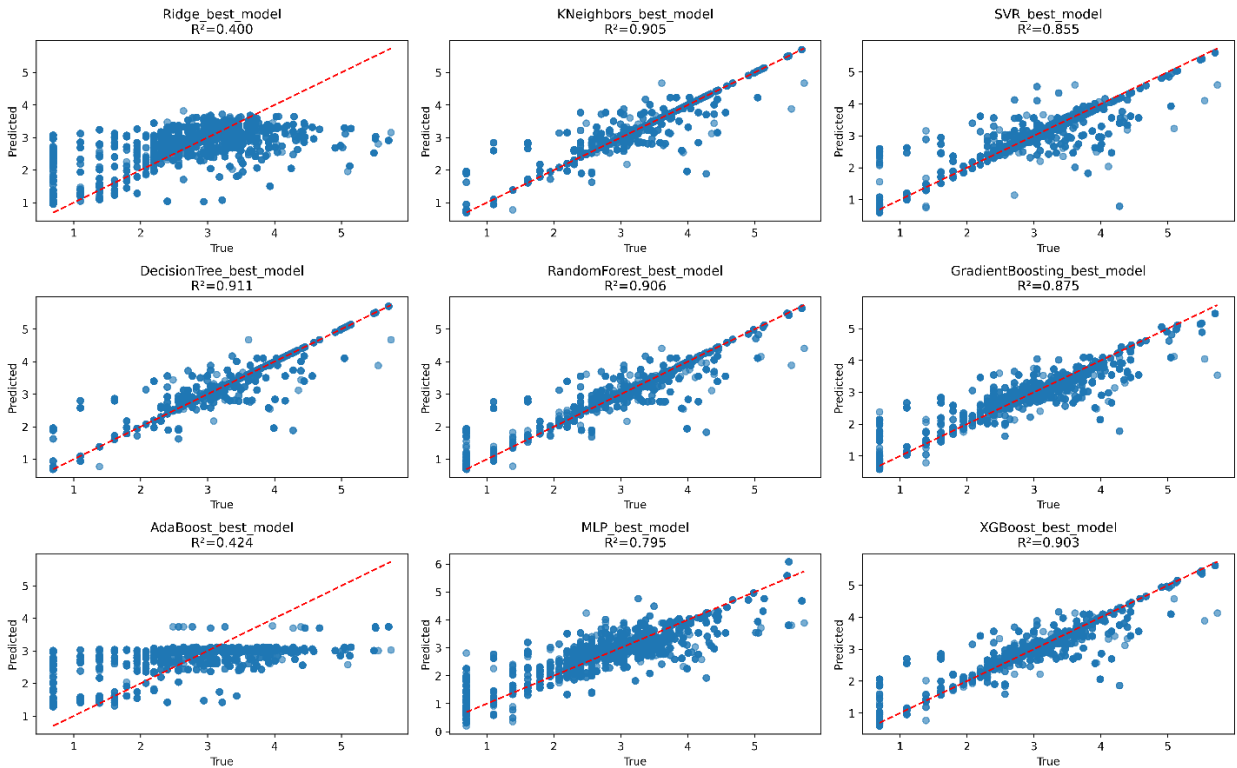

Figure 4. Predicted vs. True Value Scatter Plots for All Nine Models

## 5. Discussion

This research aimed to resolve the ambiguity surrounding the determination of qualitative sample sizes by proposing a standardized model rooted in machine learning principles. The effort to quantitatively model this complex and subjective decision represents a notable milestone for qualitative methodology, particularly since reliance solely on the subjective concept of "data saturation" compromises the scientific rigor of research. Model performance analysis indicated that the DT demonstrated the best overall performance. This finding implies that approximately 85% of the variance within the target variable is explained by the model, indicating exceptionally high predictive accuracy given the inherent complexity of the domain (Sample (Size Standard Deviation: 29.3) and the positive skewness of the distribution. This explanatory power is exceptionally high. The Stacking Regressor ensemble approach exhibited lower performance compared to the best-performing base model (DT). This suggests that the meta-learner struggled to extract substantial additional signal when the error structures of the base models were highly correlated. Quantitatively, the findings corroborate theoretical literature: the low R2 values for simple linear models (Ridge, Lasso) imply that relationships within the qualitative dataset are distinctly non-linear and rely on complex feature interactions. The model successfully demonstrates the potential to address this complexity using advanced artificial intelligence techniques.

**Determinacy of Research Design Type:** The substantial contribution of Research Design Type confirms a fundamental distinction in qualitative methodology: different designs (Phenomenology, Ethnography, Case Study) impose distinct sampling rationales. This finding validates the necessity of a design-sensitive model, as a single, universal rule is not optimal.

**The Power of Quality Emphasis (Information Power and Competence):** Information Power emerged as a strong predictor of sample size. The model's tendency to suggest a smaller sample when the participant is an expert (15 points) provides quantitative support for the concept of "information power" (Malterud et al., 2016) and aligns with theoretical assertions that quality can compensate for quantity sufficiently (Patton, 2015). Researcher Competence's high average score (18.89) indicates that researcher experience is a key factor in effective sample planning. Highly competent researchers (15 points) are better positioned to utilize smaller samples effectively, facilitating earlier saturation.

**Heterogeneity and Scope:** The Homogeneity/Heterogeneity metric, representing variance and breadth, confirmed that working with a heterogeneous participant group (high variance) necessitates a larger sample (25 points), whereas a homogeneous study suggests a smaller sample (15 points). This finding empirically supports theoretical inferences regarding the relationship between increased research scope and the necessary sample volume.

These quantitative findings successfully provide a scientific foundation for the core assumption of qualitative research: "quality, not quantity." The model takes a critical step toward objectifying qualitative methodology principles by basing a subjective decision on ten distinct measurable parameters.

Given that assertions of saturation in qualitative research are frequently unsubstantiated claims that compromise methodological credibility, this developed model aims to introduce standardization and mitigate arbitrariness in peer review processes. Researchers gain the ability to justify their sample volume rationale not merely with the vague concept of "data saturation," but with a model output that is machine learning-based and evidence-based. This substantially contributes to scientific quality and grounds qualitative research on more valid and reliable foundations.

Although this study represents a pioneering advancement in utilizing machine learning to address fundamental uncertainty in qualitative research, it is subject to certain structural limitations. The model's most significant limitation stems from the risk of subjectivity inherent in the data set creation process. The ten metrics used were scored based on expert opinions. Representing interpretive qualitative data, such as "rich information" or "deep information," solely with quantitative scores (ordinal scales) introduces a degree of unavoidable subjectivity into the metric conversion, which may impact the model's accuracy. Furthermore, the dataset's scope and diversity are constrained. Focusing on five standard designs and relying primarily on specific national and international indices (TR-Dizin, SSCI, ESCI, Scopus) may limit the model's generalizability to diverse geographical and academic traditions. Future studies should aim to mitigate these limitations to enhance the model's generalizability and reliability. Our study advances computational approaches to complexity management by integrating multivariate inputs within a dynamic prediction framework. In doing so, it operationalizes the concept of "variety engineering" (Schwaninger & Ott, 2025) and responds to recent calls for rigorously designed, systemically grounded applications (Lane, 2024). It provides a concrete example of a systemic decision-support architecture that accounts for uncertainty and feedback—where high input quality (e.g., information power) systematically reduces the required quantity (sample size), thereby embodying a key principle of system dynamics.

Future work should expand the dataset to include additional qualitative research designs (e.g., oral history, meta-synthesis) and incorporate articles from different languages to enhance the model's international generalizability and cultural context sensitivity. The data utilized spans the years 2013–2023. Future research could model sample size trends across this temporal dimension (time series data), offering insights into the evolution of methodological norms. To reduce subjectivity in the scoring process, future work should explore the utilization of Natural Language Processing (NLP) and Text Mining techniques. Automatically scoring metrics, such as Information Power or Data Quality, by extracting data directly from article abstracts or methodology sections would mitigate dependence on expert judgment and facilitate rapid expansion of the dataset. This automation is expected to strengthen the machine learning model's error decomposition and bias analysis. Future work will focus on three key advancements to enhance the model's utility and objectivity. First, we will expand the dataset to include underrepresented qualitative designs, such as oral history and meta-synthesis. Second, to directly address the limitation of subjective scoring, we will develop a Natural Language Processing (NLP) pipeline to automatically extract and score key metrics (e.g., 'Information Power', 'Data Quality') from the methodology sections of qualitative articles.

### 5.1. Practical Application and the Q-Sat AI Decision Support System

To translate the findings of this research into a tangible, accessible, and practical tool for the academic community, we have developed and deployed the Q-Sat AI (Qualitative Saturation AI), a web-based decision support system. This tool is publicly accessible and fully operational at: https://github.com/canererden/q-sat-ai. This application operationalizes the high-performing machine learning models validated in this study, allowing researchers, reviewers, and thesis advisors to input the ten key parameters (e.g., Research Scope, Information Power, Researcher Competence) discussed previously (Figure 5). In response, the system provides an immediate, evidence-based sample size estimation, along with the ensemble average from all nine trained models. The Q-Sat AI tool serves as a direct antidote to the "unsubstantiated claims" of data saturation. It provides a transparent, quantifiable, and reproducible justification for sample size selection, grounding the researcher's decision in the data-driven model rather than subjective ambiguity.

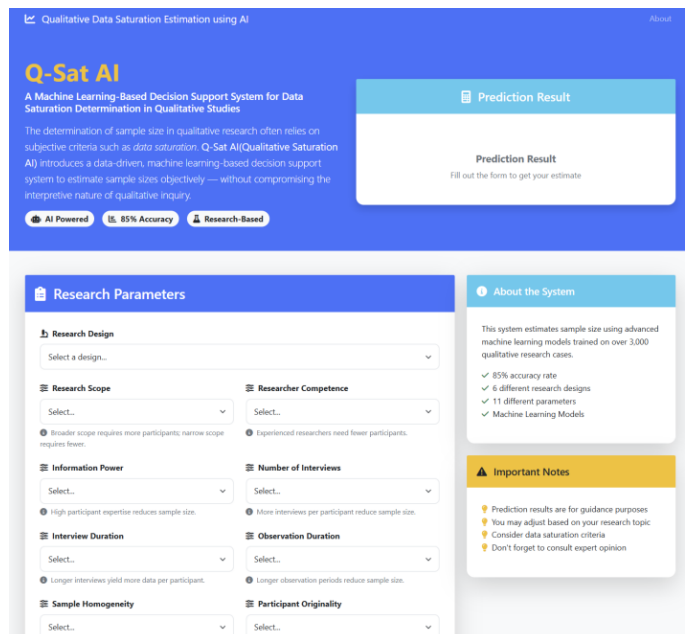

Figure 5. The web interface of the Q-Sat AI Decision Support System. Researchers input ten methodological parameters to receive an evidence-based sample size estimation.

Furthermore, this web application serves as the foundational platform for the future work outlined in our discussion. The planned integration of Conformal Prediction techniques (Angelopoulos & Bates, 2023) will enhance the tool by providing users with statistically rigorous prediction intervals, thereby quantifying the uncertainty associated with each recommendation. Similarly, the development of the NLP pipeline will serve to automate and refine the back-end scoring process, strengthening the model's objectivity. As such, Q-Sat AI moves beyond a theoretical model and functions as a live, evolving methodological instrument designed to strengthen the rigor, transparency, and scientific grounding of qualitative research.

## 6. Conclusions

This study addressed the fundamental methodological challenge of determining sample volume in qualitative research and successfully proposed an innovative, objective, and machine learning-based standard to address this pervasive uncertainty. The study demonstrated success in its primary aim of estimating the optimum sample size using qualitative research quality metrics. In the comparison among machine learning algorithms, the DT achieved the highest performance. This accuracy level is notable given the complexity and high variability inherent in qualitative research. The metrics contributing most significantly to the model's predictions were Research Design Type, Number of Participants (as an input metric), and Data Collection Method. Critical features also included Researcher Competence and Information Power. These findings quantitatively support the theoretical assumption that high quality (expertise, information power) is effective in reducing the required sample size. This study introduces a significant innovation by constructing a methodological bridge between qualitative research methodology and machine learning. The novel use of ensemble learning techniques successfully developed a more reliable and acceptable criterion without compromising the interpretive nature of qualitative research. The resulting model possesses the potential to ground future sample decisions in qualitative research on more transparent, objective, and scientific foundations. The developed computational tool is positioned as a decision support tool that rigorously *supports* human judgment. Researchers gain access to an evidence-based sample recommendation, justified by numerical metrics, thereby replacing the reliance on the vague concept of "data saturation" and mitigating a significant threat to the credibility of qualitative research. Thesis advisors and journal reviewers can utilize the model's output as a reliable reference, thereby enhancing the quality and consistency of methodological evaluations.

**Data Availability Statement**

The datasets generated and/or analyzed during the current study are available from the corresponding author upon reasonable request.

**Funding Statement**

This research was supported by the Scientific and Technological Research Council of Türkiye (TÜBİTAK) under the ARDEB program, project number 124K233.

**Conflict of Interest Disclosure**21

The authors declare that they have no known competing financial interests or personal relationships that could have appeared to influence the work reported in this paper.## References

Angelopoulos, A. N., & Bates, S. (2023). Conformal prediction: A gentle introduction. *Foundations and Trends in Machine Learning*, *16*(4), 494–591. https://doi.org/10.1561/2200000101

Awad, M., & Khanna, R. (2015). Support Vector Regression. In M. Awad & R. Khanna (Eds.), *Efficient Learning Machines: Theories, Concepts, and Applications for Engineers and System Designers* (pp. 67–80). Apress. https://doi.org/10.1007/978-1-4302-5990-9_4

Beja-Battais, P. (2023). *Overview of AdaBoost: Reconciling its views to better understand its dynamics* (No. arXiv:2310.18323). arXiv. https://doi.org/10.48550/arXiv.2310.18323

Bentéjac, C., Csörgő, A., & Martínez-Muñoz, G. (2021). A comparative analysis of gradient boosting algorithms. *Artificial Intelligence Review*, *54*(3), 1937–1967. https://doi.org/10.1007/s10462-020-09896-5

Bernard, H. R. (2000). *Social Research Methods: Qualitative and Quantitative Approaches*. SAGE.

Bernard, H. R. (2013). *Social research methods: Qualitative and quantitative approaches* (2nd ed.). Sage.

Bishop, C. M. (2006). *Pattern recognition and machine learning*. Springer.

Blockeel, H., Devos, L., Frénay, B., Nanfack, G., & Nijssen, S. (2023). Decision trees: From efficient prediction to responsible AI. *Frontiers in Artificial Intelligence*, *6*. https://doi.org/10.3389/frai.2023.1124553

Braun, V., & Clarke, V. (2021). One size fits all? What counts as quality practice in (reflexive) thematic analysis? *Qualitative Research in Psychology*, *18*(3), 328–352. https://doi.org/10.1080/14780887.2020.1769238

Breiman, L. (2001). Random Forests. *Machine Learning*, *45*(1), 5–32. https://doi.org/10.1023/A:1010933404324

Charmaz, K. (2003). Constructing grounded theory. *Qualitative Social Work*, *2*(4), 362–372.

Charmaz, K. (2006). *Constructing grounded theory: A practical guide through qualitative analysis*. SAGE.

Chen, T., & Guestrin, C. (2016). XGBoost: A Scalable Tree Boosting System. *Proceedings of the 22nd ACM SIGKDD International Conference on Knowledge Discovery and Data Mining*, 785–794. https://doi.org/10.1145/2939672.2939785

Clandinin, D. J. (2022). *Engaging in narrative inquiry* (2nd ed.). Routledge.

Clandinin, D. J., & Connelly, F. M. (2000). *Narrative inquiry: Experience and story in qualitative research*. Jossey-Bass.

Collins, G. S., Moons, K. G., Dhiman, P., Riley, R. D., Beam, A. L., Van Calster, B., Ghassemi, M., Liu, X., Reitsma, J. B., & Van Smeden, M. (2024). TRIPOD+ AI statement: Updated guidance for reporting clinical prediction models that use regression or machine learning methods. *Bmj*, *385*. https://doi.org/10.1136/bmj-2023-078378

Creswell, J. W. (1998). *Qualitative Inquiry and Research Design: Choosing Among Five Traditions*. SAGE.
22

Creswell, J. W., & Creswell, J. D. (2023). *Research design: Qualitative, quantitative, and mixed methods approaches* (6th ed.). SAGE.

Daher, W. (2023). Saturation in qualitative educational technology research. *Education Sciences*, *13*(2), 98. https://doi.org/10.3390/educsci13020098

Dawson, C. (2015). *Practical Research Methods*. Oxford.

Francis, J. J., Johnston, M., Robertson, C., Glidewell, L., Entwistle, V., Eccles, M. P., & Grimshaw, J. M. (2010). Explaining the effects of a knowledge translation intervention on professional practice: A qualitative investigation of the components of change. *Implementation Science*, *5*(1), 1–13.

Geertz, C. (1973). *The interpretation of cultures*. Basic Books.

Glaser, B. G., & Strauss, A. L. (1967). *The discovery of grounded theory: Strategies for qualitative research*. Aldine.

Guest, G., Bunce, A., & Johnson, L. (2006). How Many Interviews Are Enough?: An Experiment with Data Saturation and Variability. *Field Methods*, *18*(1), 59–82. https://doi.org/10.1177/1525822X05279903

Halder, R. K., Uddin, M. N., Uddin, Md. A., Aryal, S., & Khraisat, A. (2024). Enhancing K-nearest neighbor algorithm: A comprehensive review and performance analysis of modifications. *Journal of Big Data*, *11*(1), 113. https://doi.org/10.1186/s40537-024-00973-y

Hammersley, M., & Atkinson, P. (2019). *Ethnography: Principles in practice* (4th ed.). Routledge.

Hennink, M. M., & Kaiser, B. N. (2022). Sample sizes for saturation in qualitative research. *Social Science & Medicine*, *292*, 114523. https://doi.org/10.1016/j.socscimed.2021.114523

Kuzel, A. J. (1999). Sampling in qualitative research. In B. F. Crabtree & W. L. Miller (Eds.), *Doing qualitative research* (2nd ed., pp. 33–45). Sage.

Kvale, S. (1996). *Interviews: An introduction to qualitative research interviewing*. Sage Publications.

Lane, D. C. (2024). Engaging with diverse worldviews using system dynamics. *Systems Research and Behavioral Science*, *41*(6), 894–899. https://doi.org/10.1002/sres.3102

Laouris, Y., & Metcalf, G. (2025). Assessing the viability of virtual structured democratic dialogue. *Systems Research and Behavioral Science*, *42*(3), 587–606. https://doi.org/10.1002/sres.3006

Lincoln, Y. S., & Guba, E. G. (1985). *Naturalistic inquiry*. Sage.

Malterud, K., Siersma, V. D., & Guassora, A. D. (2016). Sample size for qualitative interview studies: Guided by information power. *Qualitative Health Research*, *26*(13), 1753–1760. https://doi.org/10.1177/1049732315617444

Marshall, B., Cardon, P., Poddar, A., & Fontenot, R. (2013). Does Sample Size Matter in Qualitative Research?: A Review of Qualitative Interviews in is Research. *Journal of Computer Information Systems*, *54*(1), 11–22. https://doi.org/10.1080/08874417.2013.11645667

Mason, M. (2010). Sample size and saturation in PhD studies using qualitative interviews. *Forum Qualitative Sozialforschung/Forum: Qualitative Social Research*, *11*(3). http://www.qualitative-research.net/index.php/fqs/article/view/1428

Mayring, P. (2000). Qualitative content analysis. *Forum: Qualitative Social Research*, *1*(2), 20.
23

Merriam, S. B. (2013). *Qualitative research: A guide to design and implementation*. Jossey-Bass.

Morse, J. M. (1994). Designing funded qualitative research. In N. K. Denzin & Y. S. Lincoln (Eds.), *Handbook of qualitative research* (pp. 220–235). Sage.

Morse, J. M. (2000). Determining Sample Size. *Qualitative Health Research*, *10*(1), 3–5. https://doi.org/10.1177/104973200129118183

Morse, J. M. (2015). Considering data saturation in qualitative research. *Qualitative Health Research*, *25*(5), 587–588.

Moustakas, C. (1994). *Phenomenological research methods*. Sage.

Onwuegbuzie, A. J., & Leech, N. L. (2007). A call for qualitative power analyses. *Qualitative and Quantitative Methods in Libraries*, *1*(3), 125–138.

O'Reilly, M., & Parker, N. (2013). "Unsatisfactory saturation": A critical exploration of the notion of saturation in qualitative research. *Qualitative Research*, *13*(2), 190–197. https://doi.org/10.1177/1468794112446106

Panda, D. K. (2024). An application of systems thinking to group-based organisations. *Systems Research and Behavioral Science*, *41*(4), 581–597. https://doi.org/10.1002/sres.2995

Patton, M. Q. (2011). *Developmental evaluation: Applying complexity concepts to enhance innovation and use*. Guilford Press.

Patton, M. Q. (2015). *Qualitative research & evaluation methods: Integrating theory and practice* (5th ed.). Sage.

Przybyła-Kasperek, M., & Marfo, K. F. (2024). A multi-layer perceptron neural network for varied conditional attributes in tabular dispersed data. *PLOS ONE*, *19*(12), e0311041. https://doi.org/10.1371/journal.pone.0311041

Rahimi, S. (2024). Saturation in qualitative research: An evolutionary concept analysis. *International Journal of Nursing Studies Advances*, *6*, 100174. https://doi.org/10.1016/j.ijnsa.2024.100174

Sandelowski, M. (1995). Sample size in qualitative research. *Research in Nursing & Health*, *18*(2), 179–183. https://doi.org/10.1002/nur.4770180211

Schwaninger, M., & Ott, S. Ch. (2025). Variety Engineering – A Cybernetic Concept with Practical Implications. In A. Quesada-Arencibia, M. Affenzeller, & R. Moreno-Díaz (Eds.), *Computer Aided Systems Theory – EUROCAST 2024* (Vol. 15173, pp. 239–257). Springer Nature Switzerland. https://doi.org/10.1007/978-3-031-82957-4_21

Shalabh. (2022). Theory of Ridge Regression Estimation with Applications. *Journal of the Royal Statistical Society Series A: Statistics in Society*, *185*(2), 742–743. https://doi.org/10.1111/rssa.12816

Smith, J. A., Flowers, P., & Larkin, M. (2022). *Interpretative phenomenological analysis* (2nd ed.). SAGE.

Straitouri, E., Wang, L., Okati, N., & Rodriguez, M. G. (2023). Improving expert predictions with conformal prediction. *International Conference on Machine Learning*, 32633–32653. https://proceedings.mlr.press/v202/straitouri23a.html

Sweeting, B., & Sutherland, S. (2023). Design's secret partner in research: Cybernetic practices for design research pedagogy. *Systems Research and Behavioral Science*, *40*(5), 765–771. https://doi.org/10.1002/sres.2974

Wutich, A., Beresford, M., & Bernard, H. R. (2024). Sample Sizes for 10 Types of Qualitative Data Analysis: An Integrative Review, Empirical Guidance, and Next Steps. *International Journal of Qualitative Methods*, *23*, 16094069241296206. https://doi.org/10.1177/16094069241296206

Yin, R. K. (2018). *Case study research and applications* (6th ed.). SAGE.

Yıldırım, A., & Şimşek, H. (2016). *Sosyal bilimlerde nitel araştırma yöntemleri*. Seçkin Yayıncılık.

Zarghami, S. A. (2024). Resilience to disruptions: A missing piece of contingency planning in projects. *International Journal of Production Research*, *62*(17), 6029–6045. https://doi.org/10.1080/00207543.2024.2306474

Zhou, X., Chen, B., Gui, Y., & Cheng, L. (2026). Conformal Prediction: A Data Perspective. *ACM Computing Surveys*, *58*(2), 1–37. https://doi.org/10.1145/3736575